# Hyperspectral Image Recovery via Hybrid Regularization

Reza Arablouei and Frank de Hoog

*Abstract*—**Natural images tend to mostly consist of smooth regions with individual pixels having highly correlated spectra. This information can be exploited to recover hyperspectral images of natural scenes from their incomplete and noisy measurements. To perform the recovery while taking full advantage of the prior knowledge, we formulate a composite cost function containing a square-error data-fitting term and *two* distinct regularization terms pertaining to spatial and spectral domains. The regularization for the spatial domain is the sum of total-variation of the image frames corresponding to all spectral bands. The regularization for the spectral domain is the $\ell_1$-norm of the coefficient matrix obtained by applying a suitable sparsifying transform to the spectra of the pixels. We use an accelerated proximal-subgradient method to minimize the formulated cost function. We analyse the performance of the proposed algorithm and prove its convergence. Numerical simulations using real hyperspectral images exhibit that the proposed algorithm offers an excellent recovery performance with a number of measurements that is only a small fraction of the hyperspectral image data size. Simulation results also show that the proposed algorithm significantly outperforms an accelerated proximal-gradient algorithm that solves the classical basis-pursuit denoising problem to recover the hyperspectral image.**

## I. INTRODUCTION

HYPERSPECTRAL imaging, also known as imaging spectroscopy, deals with the collection of electromagnetic spectral information. Hyperspectral imaging systems aim to obtain the spectrum of the radiation reflected or emitted from each pixel in the image of a scene. They realize this by acquiring radiation intensity measurements for many bands (narrow wavelength ranges) from the electromagnetic spectrum as opposed to the visual sensing of human eye that perceives the light in three visible bands of red, green, and blue. In other words, hyperspectral imaging is a process for simultaneous acquisition of spatially co-registered images in many spectrally contiguous bands. These images can be stacked into a three-dimensional structure, known as hyperspectral image datacube, for processing and analysis [1].

In 1704, Sir Isaac Newton revealed that white light could be split into several constituent colors. The subsequent advances in spectroscopy paved the way to significant discoveries in atomic and molecular physics by providing the experimental grounds [2]. Today, hyperspectral image sensing and processing systems find applications in numerous fields such as astronomy, agriculture, biomedical imaging, geosciences, mineralogy, physics, and surveillance [3], [4]. Hyperspectral images are often used to identify objects and materials or detect processes in a scene by building on the premise that certain objects/materials leave unique fingerprints in the electromagnetic spectrum. These fingerprints, known as spectral signatures, enable identification of the objects/materials that comprise the scene. For example, the spectral signatures of ferric iron minerals help mineralogists locate their deposits [5].

Typically, each pixel of a hyperspectral image covers an area containing only a few distinctive materials. Therefore, the spectrum of each pixel can be characterized as a mixture of the spectral signatures of the materials present in the area covered by the pixel. When a library of spectral signatures of the materials, which are likely to exist in a scene, is available, the spectral data of the pixels can be coded using the library endmembers. This way, a great deal of redundancy can be eliminated to ease the storing and processing burden [6], [7]. In addition, natural images at any spectral band usually encompass salient features and details that are far less voluminous compared with the raw data of radiation intensity at all pixels. This fact becomes more prominent with higher- resolution images. As a result, most natural images are highly compressible in suitable transform domains, e.g., a discrete cosine transform domain [8] or a discrete wavelet transform domain [9]. Furthermore, the information on the spatial and spectral domains have rather different natures so it is not unrealistic to assume that they can be treated separately when collecting the measurements.

There is an ever-growing interest in high-resolution hyperspectral images. This has led to the development of several hyperspectral imaging techniques that exploit the abovementioned properties pertaining to the compressibility of hyperspectral image datacubes. These so-called compressive hyperspectral imaging techniques aim to alleviate the acquisition time and sensing complexity. A mathematical tool popularly utilized for the purpose is the theory of compressive sensing [10]-[12], which relies on the assumption that

R. Arablouei and F. de Hoog are with the Commonwealth Scientific and Industrial Research Organisation, Pullenvale QLD 4069 and Acton ACT 2601, Australia (email: reza.arablouei@csiro.au, frank.dehoog@csiro.au).

hyperspectral image data are sparse or compressible in some transform domain. According to the theory of compressive sensing, incomplete and noisy measurements of a hyperspectral image, collected as random projections, can be used to reconstruct the original hyperspectral image with limited error. The reconstruction error depends on the amount of measurements as well as the noise level. Several attempts have been made to employ this theory or other helpful tools, such as adaptive direct sampling [13], to reduce the sensing complexity and capture time for hyperspectral imaging while maintaining an acceptable reconstruction performance. Among them are the works of [14]-[24].

Most of the abovementioned works use a celebrated and now-classical technique, called basis-pursuit denoising (BPDN) [25], in their reconstruction phase where the hyperspectral image is recovered from the available partial observations by minimizing a composite cost function including a square-error term and a regularization term. The regularization term is the $\ell_1$-norm of the coefficients of a three-dimensional sparsifying transform applied to the whole datacube. In this paper, we propose a hybrid regularization scheme with two distinctive terms for the spatial and spectral domains. Each regularization term is meant to promote a certain structural attribute in the hyperspectral images of natural scenes. These attributes are the paucity of abrupt variations in the spatial domain and the sparsity (compressibility) of the coefficients of an appropriate transform in the spectral domain. Some forms of hybrid regularization to simultaneously induce different types of structures have previously been studied, e.g., in [26]-[31], albeit merely in the spatial domain, i.e., only for two-dimensional image data.

We minimize the devised hybrid-regularized cost function using an accelerated proximal-subgradient algorithm. Thus, we recover a hyperspectral image from its partial and imperfect observations by finding the least-square-error fit with minimum sum of total-variations of all image frames and, at the same time, with sparsest spectra of all pixels representable in a given basis/dictionary. In doing so, we treat the spatial and spectral domains of a hyperspectral image in a distinct yet intertwined manner. We confirm the convergence of the proposed algorithm theoretically. Our numerical examinations demonstrate that the proposed algorithm offers significant improvement over an accelerated proximal-gradient algorithm that solves the relevant BPDN problem.

## II. DATA MODEL

We denote the digitized datacube of a hyperspectral image by the three-dimensional tensor $\mathbf{T} \in \mathbb{R}^{N_v \times N_h \times N_s}$ where $N_v \in \mathbb{N}$ and $N_h \in \mathbb{N}$ represent the vertical and horizontal resolution of the image, respectively, in the spatial domain and $N_s \in \mathbb{N}$ represents the spectral resolution, i.e., the number of spectral bands at each pixel. We denote the two-dimensional image frame corresponding to the $k$th ($1 \leq k \leq N_s$) spectral band, which contains the $k$th spectral element of all pixels, by $\mathbf{F}_k \in \mathbb{R}^{N_v \times N_h}$. We define the number of pixels by $N_p = N_v \times N_h$ and form the matrix $\mathbf{X} \in \mathbb{R}^{N_s \times N_p}$ as

$$\mathbf{X} = \begin{bmatrix} \text{vec}^\top\{\mathbf{F}_1\} \\ \vdots \\ \text{vec}^\top\{\mathbf{F}_{N_s}\} \end{bmatrix}$$

where $\text{vec}\{\cdot\}$ is the vectorization operator, which returns a column vector by stacking the columns of its matrix argument on top of each other, and $\text{vec}^\top\{\cdot\}$ denotes its transpose.

We assume that the spatial and spectral domains are separable in the sense that two distinct projection (multiplexing/sampling) matrices for spatial and spectral domains can be used to capture information from the hyperspectral image in a compressive fashion. This assumption simplifies the acquisition process of the hyperspectral images to a great extent. It has been shown to be a reasonable assumption in hyperspectral imaging applications [32]. We define the projection matrices of the spatial and spectral domains by $\boldsymbol{\Phi}_p \in \mathbb{R}^{M_p \times N_p}$ and $\boldsymbol{\Phi}_s \in \mathbb{R}^{M_s \times N_s}$, respectively. Here, $M_p \in \mathbb{N} \leq N_p$ and $M_s \in \mathbb{N} \leq N_s$ are the number of respective projections made independently in the spatial and spectral domains. The projected (multiplexed/sampled) data are then sensed (measured) by a detection device that may be subject to noise or sensing/measurement error. Accordingly, the incomplete and noisy matrix of acquired measurements, denoted by $\mathbf{Y} \in \mathbb{R}^{M_s \times M_p}$, is expressed as

$$\mathbf{Y} = \boldsymbol{\Phi}_s \mathbf{X} \boldsymbol{\Phi}_p^\top + \mathbf{N} \qquad (1)$$

where $\mathbf{N} \in \mathbb{R}^{M_s \times M_p}$ is the matrix of background noise/error, which is assumed to have independent normally-distributed entries with zero mean and variance $\sigma^2 \in \mathbb{R}_+$. Clearly, (1) can be written as

$$\text{vec}\{\mathbf{Y}\} = \left( \boldsymbol{\Phi}_p \otimes \boldsymbol{\Phi}_s \right) \text{vec}\{\mathbf{X}\} + \text{vec}\{\mathbf{N}\}$$

where $\otimes$ stands for the Kronecker product [33]. This expression indicates that the assumption of the inter-domain separability leads to a projection matrix for the three-dimensional hyperspectral image domain that is the Kronecker product of two projection matrices associated with the spatial and spectral domains. This Kronecker-product matrix can be described by $M_p \times N_p + M_s \times N_s$ entries as opposed to $M_p M_s \times N_p N_s$ entries without the separability assumption.

## III. BASIS PURSUIT DENOISING

Natural images can be represented by their wavelet coefficients, which are usually compressible. Moreover, spectra of natural scene pixels can often be represented by few coefficients utilizing an appropriate representation (sparsification) basis [6]. Invoking the inter-domain separability assumption, a sparse representation basis for the hyperspectral image can be devised as the Kronecker product of the representation bases for the spatial and spectral domains, which can be expressed as

$$\boldsymbol{\Psi}_p \otimes \boldsymbol{\Psi}_s = \boldsymbol{\Psi}_v \otimes \boldsymbol{\Psi}_h \otimes \boldsymbol{\Psi}_s.$$

Here, $\boldsymbol{\Psi}_p \in \mathbb{R}^{N_p \times N_p}$ is a two-dimensional discrete wavelet transform (DWT) basis matrix, i.e., $\boldsymbol{\Psi}_p = \boldsymbol{\Psi}_v \otimes \boldsymbol{\Psi}_h$ where

$\mathbf{\Psi}_v \in \mathbb{R}^{N_v \times N_v}$ and $\mathbf{\Psi}_h \in \mathbb{R}^{N_h \times N_h}$ are one-dimensional DWT basis matrices and $\mathbf{\Psi}_s \in \mathbb{R}^{N_s \times N_s}$ is an orthonormal basis matrix for sparse representation of the spectral data. Note that, in order to maintain consistency with the prevailing notation of basis matrices, we denote a DWT basis matrix such that left-multiplication of any vector by its *transpose* gives the wavelet coefficients of the vector, i.e., **B**-wavelet coefficients of a vector **a** are calculated as $\mathbf{c} = \mathbf{B}^\top \mathbf{a}$ and hence we have $\mathbf{a} = \mathbf{Bc}$.

Exploiting the abovementioned prior knowledge on sparseness of the hyperspectral image data in the basis $\mathbf{\Psi}_p \otimes \mathbf{\Psi}_s$, the hyperspectral image **X** can be recovered from the incomplete and noisy measurements **Y** by solving a basis-pursuit denoising problem (BPDN) [25]. This means that an estimate of **X** can be found as the unique solution to the following convex minimization problem:

$$\min_{\mathcal{X}} \left[ \frac{1}{2} \left\| \text{vec}\{\mathbf{Y}\} - (\mathbf{\Phi}_p \otimes \mathbf{\Phi}_s)\text{vec}\{\mathcal{X}\} \right\|_2^2 + \gamma \left\| (\mathbf{\Psi}_v \otimes \mathbf{\Psi}_h \otimes \mathbf{\Psi}_s)^\top \text{vec}\{\mathcal{X}\} \right\|_1 \right]$$

or equivalently

$$\min_{\mathcal{X}} \left[ \frac{1}{2} \left\| \mathbf{Y} - \mathbf{\Phi}_s \mathcal{X} \mathbf{\Phi}_p^\top \right\|_F^2 + \gamma \left\| \mathbf{\Psi}_s^\top \mathcal{X} \mathbf{\Psi}_p \right\|_{1,1} \right]. \quad (2)$$

Here, $\|\cdot\|_2$, $\|\cdot\|_1$, and $\|\cdot\|_F$ stand for the $\ell_2$, $\ell_1$, and Frobenius norms, respectively, and $\|\cdot\|_{1,1}$ is the entry-wise matrix $\ell_1$ norm, i.e., it returns the sum of absolute values of all entries in its matrix argument. The first term on the right-hand side of the cost function in (2) accounts for data consistency and the second term is a regularizer that promotes sparsity. The regularization parameter $\gamma \in \mathbb{R}_+$ balances a trade-off between the two terms.

An efficient approach for solving (2) is the proximal-gradient algorithm [34]-[37]. This algorithm solves the BPDN problem in an iterative manner. At each iteration, it updates the solution by taking a step along the direction opposite to the gradient of the data-consistency term followed by applying the proximity operator of the $\ell_1$-norm-based regularization term. These steps can be states as

$$\mathbf{X}_{n-1/2} = \mathbf{X}_{n-1} + \lambda_n \mathbf{\Phi}_s^\top (\mathbf{Y} - \mathbf{\Phi}_s \mathbf{X}_{n-1} \mathbf{\Phi}_p^\top) \mathbf{\Phi}_p$$

$$\mathbf{X}_n = \text{prox}_{\lambda_n w}\{\mathbf{X}_{n-1/2}\} \quad (3)$$

where we define

$$w(\mathcal{X}) = \gamma \left\| \mathbf{\Psi}_s^\top \mathcal{X} \mathbf{\Psi}_p \right\|_{1,1}. \quad (4)$$

Here, $\mathbf{X}_n$ is the estimate of **X** at iteration $n \in \mathbb{N}$ and $\lambda_n \in \mathbb{R}_+$ is the time-varying step-size. Here, $\text{prox}_f\{\mathbf{Z}\}$ denotes the proximity operator of the convex function $f$, which is defined as

$$\text{prox}_f\{\mathbf{Z}\} = \arg\min_{\mathbf{U}} \left[ f(\mathbf{U}) + \frac{1}{2} \|\mathbf{Z} - \mathbf{U}\|_F^2 \right]. \quad (5)$$

The proximity operator of the product of a weight $\xi \in \mathbb{R}$ and $\ell_1$-norm (absolute value) of a scalar $z \in \mathbb{R}$ is calculated as

$$\text{prox}_{\xi|\cdot|}\{z\} = \begin{cases} z - \xi & z > \xi \\ 0 & |z| \le \xi \\ z + \xi & z < -\xi. \end{cases} \quad (6)$$

where $|\cdot|$ is the absolute-value operator. The proximity operator of the $\xi$-weighted $\ell_1$-norm of a matrix $\mathbf{Z} \in \mathbb{R}^{N_s \times N_p}$, denoted by $\text{prox}_{\xi\|\cdot\|_{1,1}}\{\mathbf{Z}\}$, is calculated by applying (6) to all the individual entries of **Z** independently [34]. In the Appendix, we show that, in view of the orthonormality of $\mathbf{\Psi}_p$ and $\mathbf{\Psi}_s$, (3) can be written as

$$\mathbf{X}_n = \mathbf{\Psi}_s \, \text{prox}_{\lambda_n \gamma \|\cdot\|_{1,1}} \{\mathbf{\Psi}_s^\top \mathbf{X}_{n-1/2} \mathbf{\Psi}_p\} \mathbf{\Psi}_p^\top.$$

To accelerate the convergence of the above proximal-gradient algorithm, we use the acceleration scheme used in the fast iterative shrinkage/thresholding algorithm (FISTA) [38]. This scheme was originally developed in [39] to speed up the convergence of gradient-descent methods. We repeat the iterations until the stopping criterion described by

$$\|\mathbf{X}_n - \mathbf{X}_{n-1}\|_F / \|\mathbf{X}_{n-1}\|_F < \tau \text{ or } n > C \quad (7)$$

is satisfied. Here, $\tau \in \mathbb{R}_+$ is a threshold parameter and $C \in \mathbb{N}$ is the maximum allowed number of iterations. We summarize this algorithm, called accelerated proximal-gradient BPDB (APG-BPDN), in Table I.

## IV. PROPOSED ALGORITHM

The natural images tend to be constituted of mostly piece-wise smooth regions with limited rapid variations at edges of the regions. Hence, the total-variation of a natural image is typically smaller than that of its distorted or noisy versions. This property is commonly utilized to restore images with preserved edges and main features when only incomplete, noisy, or blurred versions of them are available [40]-[42]. It is also known that minimizing the total-variation of an image usually leads to a better recovery performance compared with minimizing the $\ell_1$-norm of the wavelet coefficients of the image [14], [43]. Based on this knowledge, we recover the hyperspectral data **X** from the incomplete and noisy observations **Y** by solving the following convex minimization problem:

$$\min_{\mathcal{X}} \left[ \frac{1}{2} \left\| \mathbf{Y} - \mathbf{\Phi}_s \mathcal{X} \mathbf{\Phi}_p^\top \right\|_F^2 + \gamma_1 \sum_{k=1}^{N_s} t(\mathcal{F}_k) + \gamma_2 \left\| \mathbf{\Psi}_s^\top \mathcal{X} \right\|_{1,1} \right]. \quad (8)$$

The cost function in (8) has two regularization terms. The first regularization term is the sum of total-variations of all the image frames in the hyperspectral datacube and the second regularization term is the sum of the $\ell_1$-norm of the transform coefficients of all pixels. The regularization parameters $\gamma_1 \in \mathbb{R}_+$ and $\gamma_2 \in \mathbb{R}_+$ help create a balance between explanation of the measurements and enforcement of the minimum-total-variation and sparsity features. The function $t(\mathcal{F}_k): \mathbb{R}^{N_v \times N_h} \mapsto \mathbb{R}_+$ returns the isotropic total-variation of the image frame $\mathcal{F}_k$. It is defined as

$$t(\boldsymbol{\mathcal{F}}_k) = \sum_{i=1}^{N_v} \sum_{j=1}^{N_h} \|\mathbf{d}_{i,j,k}\|_2$$

where

$$\mathbf{d}_{i,j,k} = \begin{bmatrix} d_{i,j,k}^v \\ d_{i,j,k}^h \end{bmatrix},$$

$$d_{i,j,k}^v = \begin{cases} x_{i+1,j,k} - x_{i,j,k} & i < N_v \\ 0 & i = N_v, \end{cases}$$

$$d_{i,j,k}^h = \begin{cases} x_{i,j+1,k} - x_{i,j,k} & j < N_h \\ 0 & j = N_h, \end{cases}$$

and $x_{i,j,k}$ is the $(i + (j-1)N_v, k)$th entry of $\boldsymbol{\mathcal{X}}$, which is in fact the $(i,j)$th entry of the hyperspectral image datacube corresponding to $\boldsymbol{\mathcal{X}}$ and $\boldsymbol{\mathcal{F}}_k$, $k = 1, \ldots, N_s$. Note that we use the symbols $\boldsymbol{\mathcal{X}}$ and $\boldsymbol{\mathcal{F}}_k$ as the function variables to avoid confusion with the original hyperspectral data, $\mathbf{X}$ and $\mathbf{F}_k$. However, $\boldsymbol{\mathcal{X}}$ and $\boldsymbol{\mathcal{F}}_k$, $k = 1, \ldots, N_s$, are related to each other just like $\mathbf{X}$ and $\mathbf{F}_k$, $k = 1, \ldots, N_s$, are.

Let us denote the cost function in (8) by $c(\boldsymbol{\mathcal{X}})$ and split it as

$$c(\boldsymbol{\mathcal{X}}) = a(\boldsymbol{\mathcal{X}}) + b(\boldsymbol{\mathcal{X}})$$

where

$$a(\boldsymbol{\mathcal{X}}) = \frac{1}{2}\left\|\mathbf{Y} - \boldsymbol{\Phi}_s \boldsymbol{\mathcal{X}} \boldsymbol{\Phi}_p^\top\right\|_F^2 + \gamma_1 \sum_{k=1}^{N_s} t(\boldsymbol{\mathcal{F}}_k)$$

and

$$b(\boldsymbol{\mathcal{X}}) = \gamma_2 \|\boldsymbol{\Psi}_s^\top \boldsymbol{\mathcal{X}}\|_{1,1}.$$

We solve (8) employing a proximal-*sub*gradient method that utilizes a subgradient of $a(\boldsymbol{\mathcal{X}})$ and the proximity operator of $b(\boldsymbol{\mathcal{X}})$. The iteration equations of the proposed algorithm are given by

$$\begin{aligned} \mathbf{X}_{n-1/2} &= \mathbf{X}_{n-1} \\ &+ \lambda_n \big[\boldsymbol{\Phi}_s^\top(\mathbf{Y} - \boldsymbol{\Phi}_s \mathbf{X}_{n-1}\boldsymbol{\Phi}_p^\top)\boldsymbol{\Phi}_p - \gamma_1 \mathbf{H}(\mathbf{X}_{n-1})\big] \end{aligned} \quad (9)$$

$$\mathbf{X}_n = \mathrm{prox}_{\lambda_n b}\{\mathbf{X}_{n-1/2}\} \quad (10)$$

where

$$\begin{aligned} \mathbf{H}(\mathbf{X}_{n-1}) &= \begin{bmatrix} \mathrm{vec}^\top\{\mathbf{G}(\mathbf{F}_{1,n-1})\} \\ \vdots \\ \mathrm{vec}^\top\{\mathbf{G}(\mathbf{F}_{N_s,n-1})\} \end{bmatrix} \\ &\in \partial_{\boldsymbol{\mathcal{X}}} \sum_{k=1}^{N_s} t(\mathbf{F}_{k,n-1}) \end{aligned}$$

and the function $\mathbf{G}(\mathbf{F}_{k,n-1})\colon \mathbb{R}^{N_v \times N_h} \mapsto \mathbb{R}^{N_v \times N_h}$ returns a subgradient of the total-variation of the $k$th image frame at $\mathbf{F}_{k,n-1}$, i.e., $\mathbf{G}(\mathbf{F}_{k,n-1}) \in \partial_{\boldsymbol{\mathcal{X}}} t(\mathbf{F}_{k,n-1})$. Here, $\partial_{\mathbf{A}} f(\mathbf{B})$ denotes the subdifferential (set of all subgradients) of $f$ with respect to $\mathbf{A}$ at point $\mathbf{B}$. We compute the $(i,j)$th entry of $\mathbf{G}(\boldsymbol{\mathcal{F}}_k)$ as

$$\begin{aligned} \mathscr{G}_{i,j,k} &= \begin{cases} \dfrac{d_{i-1,j,k}^v}{\|\mathbf{d}_{i-1,j,k}\|_2} & i > 1 \text{ and } \|\mathbf{d}_{i-1,j,k}\|_2 > 0 \\ 0 & i = 1 \text{ or } \|\mathbf{d}_{i-1,j,k}\|_2 = 0 \end{cases} \\ &+ \begin{cases} \dfrac{d_{i,j-1,k}^h}{\|\mathbf{d}_{i,j-1,k}\|_2} & j > 1 \text{ and } \|\mathbf{d}_{i,j-1,k}\|_2 > 0 \\ 0 & j = 1 \text{ or } \|\mathbf{d}_{i,j-1,k}\|_2 = 0 \end{cases} \\ &- \begin{cases} \dfrac{d_{i,j,k}^v}{\|\mathbf{d}_{i,j,k}\|_2} - \dfrac{d_{i,j,k}^h}{\|\mathbf{d}_{i,j,k}\|_2} & \|\mathbf{d}_{i,j,k}\|_2 > 0 \\ 0 & \|\mathbf{d}_{i,j,k}\|_2 = 0. \end{cases} \end{aligned}$$

According to the result of the Appendix, (10) can be written as

$$\mathbf{X}_n = \boldsymbol{\Psi}_s \, \mathrm{prox}_{\lambda_n \gamma_2 \|\cdot\|_{1,1}}\{\boldsymbol{\Psi}_s^\top \mathbf{X}_{n-1/2}\}. \quad (11)$$

We use the same acceleration scheme as in the APG-BPDN algorithm as well as the stopping criterion (7) in the proposed algorithm. We summarize the proposed algorithm in Table II.

### A. Non-orthonormal spectral representation matrix

For the convenience of exposition, so far, we have assumed that $\boldsymbol{\Psi}_s$ is an orthonormal basis matrix. However, if $\boldsymbol{\Psi}_s$ is not orthonormal, e.g., it is a learned dictionary, (8) can be modified as

$$\min_{\boldsymbol{\mathcal{R}}} \left[\frac{1}{2}\left\|\mathbf{Y} - \boldsymbol{\Phi}_s \boldsymbol{\Psi}_s^{-\top}\boldsymbol{\mathcal{R}}\boldsymbol{\Phi}_p^\top\right\|_F^2 + \gamma_1 \sum_{k=1}^{N_s} t(\boldsymbol{\mathcal{F}}_k) + \gamma_2 \|\boldsymbol{\mathcal{R}}\|_{1,1}\right] \quad (12)$$

by defining

$$\boldsymbol{\mathcal{R}} = \boldsymbol{\Psi}_s^\top \boldsymbol{\mathcal{X}}$$

and relating $\boldsymbol{\mathcal{F}}_k$ to $\boldsymbol{\mathcal{X}} = \boldsymbol{\Psi}_s^{-\top}\boldsymbol{\mathcal{R}}$ where $\boldsymbol{\Psi}_s^{-1}$ is the Moore-Penrose pseudoinverse of $\boldsymbol{\Psi}_s$ and $\boldsymbol{\Psi}_s^{-\top} = (\boldsymbol{\Psi}_s^{-1})^\top$.

In the same vein as described above, we can utilize the proximal-subgradient method to solve the modified optimization problem (12). The resultant iteration equations will be

$$\begin{aligned} \mathbf{R}_{n-1/2} &= \mathbf{R}_{n-1} \\ &+ \lambda_n \boldsymbol{\Psi}_s^{-1}\big[\boldsymbol{\Phi}_s^\top(\mathbf{Y} - \boldsymbol{\Phi}_s \mathbf{X}_{n-1}\boldsymbol{\Phi}_p^\top)\boldsymbol{\Phi}_p - \gamma_1 \mathbf{H}(\mathbf{X}_{n-1})\big] \end{aligned} \quad (13)$$

$$\mathbf{R}_n = \mathrm{prox}_{\lambda_n \gamma_2 \|\cdot\|_{1,1}}\{\mathbf{R}_{n-1/2}\} \quad (14)$$

$$\mathbf{X}_n = \boldsymbol{\Psi}_s^{-\top}\mathbf{R}_n \quad (15)$$

where $\mathbf{R}_n = \boldsymbol{\Psi}_s^\top \mathbf{X}_n \in \mathbb{R}^{N_s \times N_p}$ is an auxiliary matrix variable. Eliminating $\mathbf{R}_n$ from (13)-(15) gives

$$\begin{aligned} \mathbf{X}_{n-1/2} &= \mathbf{X}_{n-1} \\ &+ \lambda_n \boldsymbol{\Psi}_s^{-\top}\boldsymbol{\Psi}_s^{-1}\big[\boldsymbol{\Phi}_s^\top(\mathbf{Y} - \boldsymbol{\Phi}_s \mathbf{X}_{n-1}\boldsymbol{\Phi}_p^\top)\boldsymbol{\Phi}_p - \gamma_1 \mathbf{H}(\mathbf{X}_{n-1})\big] \\ \mathbf{X}_n &= \boldsymbol{\Psi}_s^{-\top}\mathrm{prox}_{\lambda_n \gamma_2 \|\cdot\|_{1,1}}\{\boldsymbol{\Psi}_s^\top \mathbf{X}_{n-1/2}\}. \end{aligned}$$

Therefore, when $\boldsymbol{\Psi}_s$ is not orthonormal, the subgradient on the right-hand side of (9) is left-multiplied by $\Psi_s^{-\top}\Psi_s^{-1} = (\Psi_s\Psi_s^\top)^{-1}$ and the proximal map in (11) is left-multiplied by $\Psi_s^{-\top}$ rather than $\Psi_s$.

## V. CONVERGENCE ANALYSIS

In this section, we examine the convergence of the proposed algorithm, i.e., (9) and (10), from a theoretical standpoint. Let us denote the subgradient of $a(\mathcal{X})$ at $\mathbf{X}_{n-1}$, used on the right-hand side of (9), by

$$\mathbf{S}_n = -\mathbf{\Phi}_s^\top(\mathbf{Y} - \mathbf{\Phi}_s\mathbf{X}_{n-1}\mathbf{\Phi}_p^\top)\mathbf{\Phi}_p + \gamma_1\mathbf{H}(\mathbf{X}_{n-1})$$
$$\in \partial_{\mathcal{X}}a(\mathbf{X}_{n-1}). \tag{16}$$

Substituting (16) and (9) into (10) gives

$$\mathbf{X}_n = \mathrm{prox}_{\lambda_n b}\{\mathbf{X}_{n-1} - \lambda_n\mathbf{S}_n\}$$
$$= \arg\min_{\mathcal{X}}\left[\lambda_n b(\mathcal{X}) + \frac{1}{2}\|\mathbf{X}_{n-1} - \lambda_n\mathbf{S}_n - \mathcal{X}\|_F^2\right]$$
$$= \arg\min_{\mathcal{X}}\left[b(\mathcal{X}) + \mathrm{tr}\{\mathbf{S}_n^\top\mathcal{X}\} + \frac{1}{2\lambda_n}\|\mathbf{X}_{n-1} - \mathcal{X}\|_F^2\right] \tag{17}$$

where $\mathrm{tr}\{\cdot\}$ is the matrix trace operator. An implication of (17) is that $\mathbf{0}$, the $N_s \times N_p$ zero matrix, belongs to the subdifferential of the cost function on the right-hand side of (17) at $\mathbf{X}_n$. Therefore, there exists a subgradient of $b(\mathcal{X})$ at $\mathbf{X}_n$, denoted by

$$\mathbf{P}_n \in \partial_{\mathcal{X}}b(\mathbf{X}_n),$$

such that the following equality holds:

$$\mathbf{0} = \mathbf{P}_n + \mathbf{S}_n + \frac{1}{\lambda_n}(\mathbf{X}_n - \mathbf{X}_{n-1}). \tag{18}$$

Let $\mathbf{X}_\star$ be the optimal minimizer of $c(\mathcal{X})$ and rewrite (18) as

$$\mathbf{X}_n + \lambda_n\mathbf{P}_n = \mathbf{X}_{n-1} - \lambda_n\mathbf{S}_n. \tag{19}$$

Subtracting $\mathbf{X}_\star$ from both side of (19) and calculating the square of the Frobenius norm on both sides of the resulting equation gives

$$\|\mathbf{X}_n - \mathbf{X}_\star\|_F^2 + \lambda_n^2\|\mathbf{P}_n\|_F^2 + 2\lambda_n\mathrm{tr}\{\mathbf{P}_n^\top(\mathbf{X}_n - \mathbf{X}_\star)\}$$
$$= \|\mathbf{X}_{n-1} - \mathbf{X}_\star\|_F^2 + \lambda_n^2\|\mathbf{S}_n\|_F^2 - 2\lambda_n\mathrm{tr}\{\mathbf{S}_n^\top(\mathbf{X}_{n-1} - \mathbf{X}_\star)\}$$

and subsequently

$$2\lambda_n\mathrm{tr}\{\mathbf{S}_n^\top(\mathbf{X}_{n-1} - \mathbf{X}_\star)\} + 2\lambda_n\mathrm{tr}\{\mathbf{P}_n^\top(\mathbf{X}_n - \mathbf{X}_\star)\}$$
$$= \|\mathbf{X}_{n-1} - \mathbf{X}_\star\|_F^2 - \|\mathbf{X}_n - \mathbf{X}_\star\|_F^2 + \lambda_n^2\|\mathbf{S}_n\|_F^2 \tag{20}$$
$$- \lambda_n^2\|\mathbf{P}_n\|_F^2.$$

Since $\mathbf{S}_n$ and $\mathbf{P}_n$ are subgradients of $a(\mathcal{X})$ and $b(\mathcal{X})$ at $\mathbf{X}_{n-1}$ and $\mathbf{X}_n$, respectively, we have

$$a(\mathbf{X}_{n-1}) - a(\mathbf{X}_\star) \leq \mathrm{tr}\{\mathbf{S}_n^\top(\mathbf{X}_{n-1} - \mathbf{X}_\star)\} \tag{21}$$

and

$$b(\mathbf{X}_n) - b(\mathbf{X}_\star) \leq \mathrm{tr}\{\mathbf{P}_n^\top(\mathbf{X}_n - \mathbf{X}_\star)\}. \tag{22}$$

From (20)-(22), we get

$$2\lambda_n[a(\mathbf{X}_{n-1}) + b(\mathbf{X}_n) - a(\mathbf{X}_\star) - b(\mathbf{X}_\star)]$$
$$\leq \|\mathbf{X}_{n-1} - \mathbf{X}_\star\|_F^2 - \|\mathbf{X}_n - \mathbf{X}_\star\|_F^2 + \lambda_n^2\|\mathbf{S}_n\|_F^2 \tag{23}$$
$$- \lambda_n^2\|\mathbf{P}_n\|_F^2.$$

Replacing the iteration index $n$ with $m$ in (23) then summing up both sides for $m = 1, \ldots, n$, results in

$$2\sum_{m=1}^n \lambda_m[a(\mathbf{X}_{m-1}) + b(\mathbf{X}_m) - c(\mathbf{X}_\star)]$$
$$\leq \|\mathbf{X}_0 - \mathbf{X}_\star\|_F^2 - \|\mathbf{X}_n - \mathbf{X}_\star\|_F^2 + \sum_{m=1}^n \lambda_m^2\|\mathbf{S}_m\|_F^2$$
$$- \sum_{m=1}^n \lambda_m^2\|\mathbf{P}_m\|_F^2$$
$$\leq \|\mathbf{X}_0 - \mathbf{X}_\star\|_F^2 + \sum_{m=1}^n \lambda_m^2\|\mathbf{S}_m\|_F^2 \tag{24}$$

where $\mathbf{X}_0$ is an arbitrary finite-valued matrix chosen as the initial estimate of $\mathbf{X}$.

Assuming a fixed step-size, i.e., $\lambda_n = \lambda\ \forall n$, (24) leads to

$$\sum_{m=0}^{n-1}[c(\mathbf{X}_m) - c(\mathbf{X}_\star)]$$
$$\leq \frac{1}{2\lambda}\|\mathbf{X}_0 - \mathbf{X}_\star\|_F^2 + \frac{\lambda}{2}\sum_{m=1}^n\|\mathbf{S}_m\|_F^2 + b(\mathbf{X}_0) - b(\mathbf{X}_n)$$
$$\leq \frac{1}{2\lambda}\|\mathbf{X}_0 - \mathbf{X}_\star\|_F^2 + \frac{\lambda}{2}\sum_{m=1}^n\|\mathbf{S}_m\|_F^2 + b(\mathbf{X}_0) \tag{25}$$

where we factor in that the function $b(\mathcal{X})$ is nonnegative for any value of $\mathcal{X}$, i.e.,

$$b(\mathcal{X}) \geq 0, \forall \mathcal{X}.$$

Denote the best (smallest) value of the cost function $c(\mathcal{X})$ attained over $n-1$ iterations as $\check{c}_{n-1}$. It is clear that

$$\check{c}_{n-1} - c(\mathbf{X}_\star) \leq \frac{1}{n}\sum_{m=0}^{n-1}[c(\mathbf{X}_m) - c(\mathbf{X}_\star)]. \tag{26}$$

From (25) and (26), we obtain

$$\check{c}_{n-1} - c(\mathbf{X}_\star)$$
$$\leq \frac{1}{2n\lambda}\|\mathbf{X}_0 - \mathbf{X}_\star\|_F^2 + \frac{\lambda}{2n}\sum_{m=1}^n\|\mathbf{S}_m\|_F^2 + \frac{1}{n}b(\mathbf{X}_0). \tag{27}$$

If there exists a finite nonnegative constant $\zeta \in \mathbb{R}_+$ that fulfils

$$\|\mathbf{S}_m\|_F \leq \zeta \text{ for } m = 1, \ldots, n-1, \tag{28}$$

(27) will turn into

$$\check{c}_{n-1} - c(\mathbf{X}_\star) \leq \frac{1}{2n\lambda}\|\mathbf{X}_0 - \mathbf{X}_\star\|_F^2 + \frac{\lambda}{2}\zeta^2 + \frac{1}{n}b(\mathbf{X}_0). \tag{29}$$

For a sufficiently large number of iterations, i.e., $n \to \infty$, (29) gives

$$\check{c}_{n-1} - c(\mathbf{X}_\star) \leq \frac{\lambda}{2}\zeta^2. \tag{30}$$

The inequality (30) indicates that the proximal-subgradient algorithm of (9) and (11) with a constant step-size $\lambda_n = \lambda$ converges to the vicinity of the optimal solution where the error in the achieved value of the cost function is not greater than $\lambda \zeta^2 / 2$. It can be expected that the error will vanish asymptotically when a variable step-size that diminishes in iterations is used.

Our analysis essentially shows that $b(\mathcal{X})$ does not harm the convergence over what is achievable when minimizing $a(\mathcal{X})$. The total-variation function $\sum_{k=1}^{N_s} t(\mathcal{F}_k)$ and its subgradient $\mathbf{H}(\mathcal{X})$ are highly nonlinear. Therefore, it is difficult to evaluate a value of $\zeta$ that satisfies (28). However, since $\sum_{k=1}^{N_s} t(\mathcal{F}_k)$ and consequently $a(\mathcal{X})$ are convex, when the step-size is chosen such that the subgradient algorithm minimizing $a(\mathcal{X})$ via $\mathbf{S}_n$ converges, $\mathbf{X}_m$, $m = 1, \ldots, n$, remain finite and there exists a finite $\zeta$ to make (28) hold, hence the proximal-subgradient algorithm converges.

Our analysis applies to the non-accelerated version of the proposed algorithm, given in Table II. However, it can be extended to cover the accelerated version following an approach similar to the one taken in [38]. Moreover, it is evidently straightforward to deduce the same theoretical results for the algorithm described by (13)-(15), which relates to the case of having a non-orthonormal $\boldsymbol{\Psi}_s$.

In general, subgradient algorithms converge slowly, i.e., with a guaranteed rate of $\mathcal{O}(1/\sqrt{n})$, to the best achievable solution [44]. Nonetheless, as we will see in the next section (Fig. 4 ahead), convergence speed of the proposed algorithm is similar to that of the APG-BPDN algorithm, which is based on an accelerated gradient-descent method. The function $a(\mathcal{X})$ is not strictly smooth or differentiable at all points due to the total-variation regularization term $\sum_{k=1}^{N_s} t(\mathcal{F}_k)$. However, the quadratic data-fidelity-enforcing term $\frac{1}{2} \left\| \mathbf{Y} - \boldsymbol{\Phi}_s \mathcal{X} \boldsymbol{\Phi}_p^{\top} \right\|_F^2$ is smooth and normally has a significant weight in $a(\mathcal{X})$. Therefore, in practice, $a(\mathcal{X})$ is rather smooth and the convergence of the proposed algorithm is almost linear in time (iterations). This can also justify why the stopping criterion (7) works well for the proposed algorithm despite the fact that determining a proper stopping criterion for subgradient algorithms is usually non-trivial.

## VI. Simulation Results

We consider six hyperspectral images, titled Stanford Dish, San Francisco, Harvard Outdoor, Harvard Indoor, Indian Pines, and Washington DC Mall (see Fig. 4 for their RGB renderings). The Stanford Dish and San Francisco images are from the Stanford Center for Image Systems Engineering [45]. The Harvard Outdoor and Harvard Indoor images are from the Computer Vision Laboratory of the Harvard University [46]. The Washington DC Mall and Indian Pines images are from the Purdue University Research Repository [47], [48]. We resize the original images such that they have spatial and spectral resolutions as given in Table III.

Natural images are known to have most of their energy concentrated in the lower parts of their two-dimensional Fourier or Walsh spectra [49]-[51]. Therefore, we use a multiplexing matrix for the spatial domain that is composed of two parts, i.e.,

$$\boldsymbol{\Phi}_p = \begin{bmatrix} \boldsymbol{\Phi}_{p,1} \\ \boldsymbol{\Phi}_{p,2} \end{bmatrix}.$$

The first part, $\boldsymbol{\Phi}_{p,1} \in \mathbb{R}^{Q_p \times N_p}$, extracts $Q_p \in \mathbb{N}$ top-left coefficients of the two-dimensional Walsh-Hadamard transform (WHT) of the vectorized image that it multiplies. The coefficients are chosen in an order that is similar to the zig-zagging pattern used by the JPEG still-image data compression standard [8]. The entries of the second part, $\boldsymbol{\Phi}_{p,2} \in \mathbb{R}^{(M_p - Q_p) \times N_p}$, take random values drawn from a Rademacher (symmetric Bernoulli) distribution to implement $M_p - Q_p$ random projections. Therefore, we have

$$\mathbf{X} \boldsymbol{\Phi}_p^{\top} = \begin{bmatrix} \text{vec}^{\top}\{\mathbf{F}_1\} \\ \vdots \\ \text{vec}^{\top}\{\mathbf{F}_{N_s}\} \end{bmatrix} \begin{bmatrix} \boldsymbol{\Phi}_{p,1}^{\top}, \boldsymbol{\Phi}_{p,2}^{\top} \end{bmatrix}$$

$$= \begin{bmatrix} (\boldsymbol{\Phi}_{p,1}\text{vec}\{\mathbf{F}_1\})^{\top}, (\boldsymbol{\Phi}_{p,2}\text{vec}\{\mathbf{F}_1\})^{\top} \\ \vdots \\ (\boldsymbol{\Phi}_{p,1}\text{vec}\{\mathbf{F}_{N_s}\})^{\top}, (\boldsymbol{\Phi}_{p,2}\text{vec}\{\mathbf{F}_{N_s}\})^{\top} \end{bmatrix}$$

$$= \begin{bmatrix} \left\{ \mathbf{q}_{Q_p}\left(\mathbf{W}_{N_v}^{\top}\mathbf{F}_1\mathbf{W}_{N_h}\right) \right\}^{\top}, (\boldsymbol{\Phi}_{p,2}\text{vec}\{\mathbf{F}_1\})^{\top} \\ \vdots \\ \left\{ \mathbf{q}_{Q_p}\left(\mathbf{W}_{N_v}^{\top}\mathbf{F}_{N_s}\mathbf{W}_{N_h}\right) \right\}^{\top}, (\boldsymbol{\Phi}_{p,2}\text{vec}\{\mathbf{F}_1\})^{\top} \end{bmatrix}$$

where $\mathbf{W}_{N_v} \in \mathbb{R}^{N_v \times N_v}$ and $\mathbf{W}_{N_h} \in \mathbb{R}^{N_h \times N_h}$ are the sequency-ordered WHT basis matrices of order $N_v$ and $N_h$, respectively, and the function $\mathbf{q}_B(\mathbf{A}): \mathbb{R}^{A_1 \times A_2} \mapsto \mathbb{R}^{B \times 1}$ returns a column vector containing $B$ top-left entries of $\mathbf{A}$ picked in the order as shown in Fig. 1.

We use a similar approach to construct the multiplexing matrix of the spectral domain, i.e., we arrange

$$\boldsymbol{\Phi}_s = \begin{bmatrix} \boldsymbol{\Phi}_{s,1} \\ \boldsymbol{\Phi}_{s,2} \end{bmatrix}$$

where $\boldsymbol{\Phi}_{s,1} \in \mathbb{R}^{Q_s \times N_s}$ contains $Q_s \in \mathbb{N}$ top rows of the sequency-ordered Hadamard matrix of order $N_s$ and $\boldsymbol{\Phi}_{s,2} \in \mathbb{R}^{(M_s - Q_s) \times N_s}$ consists of random-valued entries with Rademacher distribution. Therefore, if we express $\mathbf{X}$ as

$$\mathbf{X} = \begin{bmatrix} \mathbf{x}_1, \ldots, \mathbf{x}_{N_p} \end{bmatrix}$$

where $\mathbf{x}_l \in \mathbb{R}^{N_s \times 1}$, $1 \leq l \leq N_p$, corresponds to the spectrum of $l$th pixel, we can write

$$\boldsymbol{\Phi}_s \mathbf{X} = \begin{bmatrix} \boldsymbol{\Phi}_{s,1}\mathbf{x}_1 & \boldsymbol{\Phi}_{s,1}\mathbf{x}_{N_p} \\ \boldsymbol{\Phi}_{s,2}\mathbf{x}_1 & \cdots & \boldsymbol{\Phi}_{s,2}\mathbf{x}_{N_p} \end{bmatrix}$$

$$= \begin{bmatrix} \mathbf{q}_{Q_s}\left(\mathbf{W}_{N_s}^{\top}\mathbf{x}_1\right), & \mathbf{q}_{Q_s}\left(\mathbf{W}_{N_s}^{\top}\mathbf{x}_{N_p}\right) \\ \boldsymbol{\Phi}_{s,2}\mathbf{x}_1 & , \ldots, & \boldsymbol{\Phi}_{s,2}\mathbf{x}_{N_p} \end{bmatrix}.$$

The main advantage of using the above multiplexing matrices is that their entries only take values of $\pm 1$ and can be realized via the fast WHT requiring minimal memory and computations [52]. This makes them suitable for real-world applications where they can be implemented using digital micro-mirror devices (DMDs) or multiplexed sensor arrays, e.g., as depicted in Fig. 3 of [14].

We use the two-dimensional Haar wavelet basis matrix as the representation basis for the spatial domain, $\mathbf{\Psi}_p$, in the APG-BPDN algorithm. In order to find a good representation basis matrix for the spectral domain, $\mathbf{\Psi}_s$, we randomly select one percent of the pixels, i.e., $\hat{N}_p = N_p/100$ pixels, and arrange their spectra in a matrix called $\hat{\mathbf{X}} \in \mathbb{R}^{N_s \times \hat{N}_p}$. We then perform an economical singular-value decomposition of $\hat{\mathbf{X}}$ to obtain

$$\hat{\mathbf{X}} = \acute{\mathbf{U}} \acute{\mathbf{\Sigma}} \acute{\mathbf{V}}^\top$$

and use the calculated left-singular-vector subspace $\acute{\mathbf{U}}$ as the spectral basis matrix, i.e., $\mathbf{\Psi}_s = \acute{\mathbf{U}}$.

We assume that the measurements, $\mathbf{\Phi}_s \mathbf{X} \mathbf{\Phi}_p^\top$, are contaminated with additive zero-mean Gaussian noise of standard deviation $\sigma = 10^{-2}$. We set the number of low-pass measurements, $Q_p$ and $Q_s$, to approximately ten and five percent of the number of pixels and spectral bands, respectively, i.e., $Q_p = \lceil 0.1 \times N_p \rceil$ and $Q_s = \lceil 0.05 \times N_s \rceil$ where the operator $\lceil \cdot \rceil$ rounds its argument to the nearest integer. The value of $Q_p$ and $Q_s$ for the considered hyperspectral images are shown in Table III.

We recover the considered hyperspectral images from their noisy partial measurements using the APG-BPDN and the proposed algorithms. We also use the TVAL3 algorithm of [53] when the measurements are complete in the spectral domain, i.e., when $M_s = N_s$.

The TVAL3 algorithm is one of the most popular algorithms for recovering two-dimensional images from their compressive measurements through total-variation minimization. It utilizes the augmented Lagrangian method together with an alternating-direction approach and backtracking line-search. Since the TVAL3 algorithm is originally designed for recovering two-dimensional images, we apply it to all the frames of the considered hyperspectral images when no projection/multiplexing is implemented in the spectral domain, i.e., when $\mathbf{\Phi}_s = \mathbf{I}_{N_s}$. In our experiments with the TVAL3 algorithm, we use it in the isotropic TV/L2 mode and tune its parameters according to the instructions given in [54] to achieve the best possible performance for each considered hyperspectral image.

We initialize the APG-BPDN and the proposed algorithms with $\mathbf{X}_0 = \mathbf{\Phi}_s^\top \mathbf{Y} \mathbf{\Phi}_p$, use a fixed step-size $\lambda_n = 0.25$, and set the threshold for stopping criterion and the number of maximum iterations to $\tau = 10^{-3}$ and $C = 200$, respectively. We have determined the best values for the regularization parameters $\gamma_1$ and $\gamma_2$ in the proposed algorithm and $\gamma$ in the APG-BPDN algorithm through extensive experimentations with each considered hyperspectral image. We list these values in Table III. Our experiments show that these parameter values yield almost the best achievable performance for both algorithms in the considered scenarios.

As the performance measure, we use the relative error, defined by

$$\frac{\|\mathbf{X} - \mathbf{X}_\infty\|_F^2}{\|\mathbf{X}\|_F^2}$$

where $\mathbf{X}_\infty$ denotes the recovered hyperspectral image that is the converged value of $\mathbf{X}_n$.

In Fig. 2, we plot the relative errors of recovering the considered hyperspectral images via the APG-BPDN, the TVAL3, and the proposed algorithms as functions of the spatial measurement rate, called $r_p$, for different values of the spectral measurement rate, called $r_s$. For any given $r_p$ and $r_s$, we calculate the number of spatial and spectral projections as $M_p = \lfloor r_p N_p \rfloor$ and $M_s = \lceil r_s N_s \rceil$. The TVAL3 algorithm is applicable only when $r_s = 1$.

In Fig. 3, we plot the convergence curves of the APG-BPDN and the proposed algorithms as the evolution of the relative error in iterations when recovering the considered hyperspectral images with $r_p = 0.2$, $r_s = 0.1$ and $r_p = 0.5$, $r_s = 0.2$.

Figs. 2 and 3 demonstrate that the proposed algorithm has a good performance and significantly outperforms its conventional counterpart, the APG-BPDN algorithm. Fig 2 also shows that the performance of the proposed algorithm is superior to that of the TVAL3 algorithm when no compression is applied in the spectral domain.

In Table IV, we provide the number of iterations as well as the processing time required by each algorithm to converge. The results are for different values of $r_p$ and $r_s$. To facilitate the comparison, we also include the relative error for each case. We implement the simulations using MATLAB on a Mobile Workstation with a 2.9GHz Core-i7 CPU and 24GB of DDR3 RAM.

According to the runtimes in Table IV, the proposed algorithm is appreciably faster than the APG-BPDN and the TVAL3 algorithms. Note that the TVAL3 algorithm is inherently restricted to handle only two-dimensional images in the spatial domain. Therefore, the iteration numbers given in Table IV are averaged over $N_s$ runs of this algorithm in each experiment where each run corresponds to one frame of the hyperspectral image to be recovered. In addition, as the iterative optimization technique used in the TVAL3 algorithm is of a very different nature compared to the ones used in the APG-BPDN and the proposed algorithms, comparing the number of iterations of the TVAL3 algorithm with those of the APG-BPDN and the proposed algorithms would not be informative. We include them in Table IV only for the sake of completeness.

Fig. 4 contains the RGB illustrations of the considered hyperspectral images and their reconstructed versions using the APG-BPDN and the proposed algorithms when $r_p = 0.2$ and $r_s = 0.1$. Figs. 5 and 6 contain the RGB illustrations when $r_p = 0.2$ and $r_s = 1$ and include the images reconstructed by the TVAL3 algorithm. Images in Fig. 6 are zoomed to the bottom left quarter. In Figs. 4-6, the difference between the original image and the one recovered by the proposed algorithm is barely noticeable for most considered hyperspectral images.

In Fig. 7, we depict the original as well as reconstructed spectra of four pixels randomly selected from the Stanford Dish and San Francisco hyperspectral images using the proposed algorithm when $r_p = 0.2$ and $r_s = 0.1$. It is seen that the reconstructed spectra match the original ones well even for very small number of measurements. With $r_p = 0.2$ and $r_s = 0.1$, the number of measurements is only 3% of the size of the hyperspectral datacube. This includes the initial 1% used for the calculation of $\boldsymbol{\Psi}_s$.

## VII. Conclusion

We studied the recovery of a hyperspectral image from its incomplete and noisy measurements. We proposed a cost function comprising two distinct regularization terms corresponding to the spatial and spectral domains. This hybrid regularization scheme allows us to minimize the total-variation of the image frames in the spatial domain and promote sparsity of the pixel spectra in the spectral domain simultaneously. We used an accelerated proximal-subgradient algorithm for minimizing the devised cost function. We proved the convergence of the proposed algorithm analytically. Simulation results corroborate the good performance of the proposed algorithm as well as its superiority over an accelerated proximal-gradient algorithm that solves the pertinent basis-pursuit denoising problem.

## Appendix

Let $\mathbf{A}$ and $\mathbf{B}$ be orthonormal matrices and

$$g(\mathbf{Z}) = \xi f(\mathbf{A}^\top \mathbf{Z} \mathbf{B}). \tag{31}$$

From (31) and the definition of the proximity operator (5), one can infer that

$$
\begin{aligned}
\mathbf{U} &= \mathrm{prox}_g\{\mathbf{Z}\} \\
&\Leftrightarrow \mathbf{0} \in \partial_{\mathbf{U}} \left( \xi f(\mathbf{A}^\top \mathbf{U} \mathbf{B}) + \frac{1}{2} \|\mathbf{Z} - \mathbf{U}\|_F^2 \right) \\
&\Leftrightarrow \mathbf{0} \in \xi \, \partial_{\mathbf{U}} f(\mathbf{A}^\top \mathbf{U} \mathbf{B}) + \mathbf{U} - \mathbf{Z}
\end{aligned} \tag{32}
$$

where $\partial_{\mathbf{U}} f(\mathbf{V})$ denotes the subdifferential (set of all subgradients) of function $f$ with respect to $\mathbf{U}$ at point $\mathbf{V}$. It is easy to verify from (32) that

$$\mathbf{0} \in \xi \mathbf{A}[\partial_{\mathbf{A}^\top \mathbf{U} \mathbf{B}} f(\mathbf{A}^\top \mathbf{U} \mathbf{B})]\mathbf{B}^\top + \mathbf{U} - \mathbf{Z}. \tag{33}$$

Since $\mathbf{A}$ and $\mathbf{B}$ are orthonormal, (33) implies that

$$
\begin{aligned}
&\mathbf{0} \in \xi \, \partial_{\mathbf{A}^\top \mathbf{U} \mathbf{B}} f(\mathbf{A}^\top \mathbf{U} \mathbf{B}) + \mathbf{A}^\top \mathbf{U} \mathbf{B} - \mathbf{A}^\top \mathbf{Z} \mathbf{B} \\
&\Leftrightarrow \mathbf{A}^\top \mathbf{U} \mathbf{B} = \mathrm{prox}_{\xi f}\{\mathbf{A}^\top \mathbf{Z} \mathbf{B}\}.
\end{aligned}
$$

Subsequently, we have

$$\mathbf{U} = \mathbf{A} \, \mathrm{prox}_{\xi f}\{\mathbf{A}^\top \mathbf{Z} \mathbf{B}\} \mathbf{B}^\top$$

or

$$\mathrm{prox}_g\{\mathbf{Z}\} = \mathbf{A} \, \mathrm{prox}_{\xi f}\{\mathbf{A}^\top \mathbf{Z} \mathbf{B}\} \mathbf{B}^\top.$$

## Acknowledgment

We would like to thank our colleagues in CSIRO for their supports, thoughts, and discussions that contributed to improving the manuscript.

TABLE I
THE APG-BPDN ALGORITHM

initialize

$\mathbf{X}_0 = \mathbf{\Phi}_s^\top \mathbf{Y} \mathbf{\Phi}_p, \bar{\mathbf{X}}_0 = \mathbf{X}_0$

$n = 0, \alpha_0 = 1$

repeat

$n = n + 1$

$\bar{\mathbf{X}}_{n-1/2} = \mathbf{X}_{n-1} + \lambda_n \mathbf{\Phi}_s^\top (\mathbf{Y} - \mathbf{\Phi}_s \mathbf{X}_{n-1} \mathbf{\Phi}_p^\top) \mathbf{\Phi}_p$

$\bar{\mathbf{X}}_n = \mathbf{\Psi}_s \, \text{prox}_{\lambda_n \gamma \|\cdot\|_{1,1}} \{\mathbf{\Psi}_s^\top \bar{\mathbf{X}}_{n-1/2} \mathbf{\Psi}_p\} \mathbf{\Psi}_p^\top$

$\alpha_n = \dfrac{1 + \sqrt{1 + 4\alpha_{n-1}^2}}{2}$

$\mathbf{X}_n = \bar{\mathbf{X}}_n + \dfrac{\alpha_{n-1} - 1}{\alpha_n} (\bar{\mathbf{X}}_n - \bar{\mathbf{X}}_{n-1})$

stop if $\|\mathbf{X}_n - \mathbf{X}_{n-1}\|_F / \|\mathbf{X}_{n-1}\|_F < \tau$ or $n > C$

TABLE II
THE PROPOSED ALGORITHM

initialize

$\mathbf{X}_0 = \mathbf{\Phi}_s^\top \mathbf{Y} \mathbf{\Phi}_p, \bar{\mathbf{X}}_0 = \mathbf{X}_0$

$n = 0, \alpha_0 = 1$

repeat

$n = n + 1$

$\bar{\mathbf{X}}_{n-1/2} = \mathbf{X}_{n-1} + \lambda_n [\mathbf{\Phi}_s^\top (\mathbf{Y} - \mathbf{\Phi}_s \mathbf{X}_{n-1} \mathbf{\Phi}_p^\top) \mathbf{\Phi}_p - \gamma_1 \mathbf{H}(\mathbf{X}_{n-1})]$

$\bar{\mathbf{X}}_n = \mathbf{\Psi}_s \, \text{prox}_{\lambda_n \gamma_2 \|\cdot\|_{1,1}} \{\mathbf{\Psi}_s^\top \bar{\mathbf{X}}_{n-1/2}\}$

$\alpha_n = \dfrac{1 + \sqrt{1 + 4\alpha_{n-1}^2}}{2}$

$\mathbf{X}_n = \bar{\mathbf{X}}_n + \dfrac{\alpha_{n-1} - 1}{\alpha_n} (\bar{\mathbf{X}}_n - \bar{\mathbf{X}}_{n-1})$

stop if $\|\mathbf{X}_n - \mathbf{X}_{n-1}\|_F / \|\mathbf{X}_{n-1}\|_F < \tau$ or $n > C$

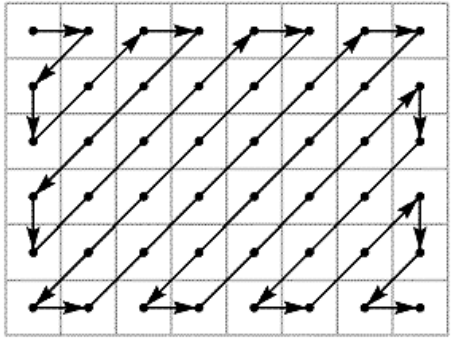

Fig. 1. Zig-zag scanning pattern used for selecting the coefficients of the Walsh-Hadamard transform. This is an example for when $N_v = 6$ and $N_h = 8$. This pattern is similar to the one used in the JPEG still-image compression standard.

TABLE III
THE CONSIDERED HYPERSPECTRAL IMAGES AND THE ASSOCIATED PARAMETER VALUES

| hyperspectral image | reference | $N_v \times N_h$ (spatial resolution) | $N_s$ (spectral bands) | $Q_p$ | $Q_s$ | $\gamma$ | $\gamma_1$ | $\gamma_2$ |
|---|---|---|---|---|---|---|---|---|
| Stanford Dish and San Francisco | [45] | $512 \times 512$ | 128 | 26214 | 6 | 0.1 | $2 \times 10^{-4}$ | $2 \times 10^{-3}$ |
| Harvard Outdoor and Indoor | [46] | $1024 \times 1024$ | 32 | 104858 | 2 | 0.3 | $1 \times 10^{-3}$ | $4 \times 10^{-3}$ |
| Washington DC Mall | [47] | $256 \times 256$ | 128 | 6554 | 6 | 0.05 | $2 \times 10^{-3}$ | $4 \times 10^{-3}$ |
| Indian Pines | [48] | $512 \times 512$ | 64 | 26214 | 3 | 0.15 | $1 \times 10^{-3}$ | $2 \times 10^{-3}$ |

TABLE IV
RELATIVE ERROR OF THE APG-BPDN, THE TVAL3, AND THE PROPOSED ALGORITHMS AS WELL AS THE NUMBER OF ITERATIONS AND PROCESSING TIME
REQUIRED FOR THEIR CONVERGENCE WITH DIFFERENT VALUES OF SPATIAL AND SPECTRAL MEASUREMENT RATES

(A) STANFORD DISH

| algorithm | $r_p$ | $r_s$ | relative error | iterations | time |
|---|---|---|---|---|---|
| APG-BPDN | 0.2 | 0.1 | 9.17% | 169 | 36 min 29 sec |
| proposed | 0.2 | 0.1 | 7.00% | 115 | 22 min 31 sec |
| APG-BPDN | 0.2 | 1 | 8.74% | 87 | 16 min 9 sec |
| TVAL3 | 0.2 | 1 | 7.01% | 32 | 10 min 39 sec |
| proposed | 0.2 | 1 | 6.75% | 76 | 9 min 54 sec |
| APG-BPDN | 0.5 | 0.2 | 6.62% | 145 | 35 min 19 sec |
| proposed | 0.5 | 0.2 | 4.23% | 90 | 19 min 54 sec |
| APG-BPDN | 0.5 | 1 | 5.91% | 74 | 14 min 20 sec |
| TVAL3 | 0.5 | 1 | 4.21% | 34 | 11 min 40 sec |
| proposed | 0.5 | 1 | 3.90% | 64 | 9 min 34 sec |

(B) SAN FRANCISCO

| algorithm | $r_p$ | $r_s$ | relative error | iterations | time |
|---|---|---|---|---|---|
| APG-BPDN | 0.2 | 0.1 | 9.92% | 178 | 38 min 8 sec |
| proposed | 0.2 | 0.1 | 7.03% | 120 | 23 min 0 sec |
| APG-BPDN | 0.2 | 1 | 9.47% | 98 | 18 min 21 sec |
| TVAL3 | 0.2 | 1 | 7.00% | 32 | 10 min 39 sec |
| proposed | 0.2 | 1 | 6.68% | 81 | 10 min 12 sec |
| APG-BPDN | 0.5 | 0.2 | 7.21% | 156 | 38 min 19 sec |
| proposed | 0.5 | 0.2 | 4.17% | 89 | 20 min 2 sec |
| APG-BPDN | 0.5 | 1 | 6.54% | 76 | 14 min 34 sec |
| TVAL3 | 0.5 | 1 | 4.13% | 37 | 11 min 51 sec |
| proposed | 0.5 | 1 | 3.81% | 65 | 10 min 8 sec |

(C) HARVARD OUTDOOR

| algorithm | $r_p$ | $r_s$ | relative error | iterations | time |
|---|---|---|---|---|---|
| APG-BPDN | 0.2 | 0.1 | 9.82% | 198 | 80 min 10 sec |
| proposed | 0.2 | 0.1 | 5.88% | 113 | 42 min 19 sec |
| APG-BPDN | 0.2 | 1 | 8.48% | 70 | 16 min 14 sec |
| TVAL3 | 0.2 | 1 | 4.21% | 28 | 10 min 39 sec |
| proposed | 0.2 | 1 | 4.15% | 62 | 10 min 7 sec |
| APG-BPDN | 0.5 | 0.2 | 8.42% | 168 | 84 min 20 sec |
| proposed | 0.5 | 0.2 | 4.29% | 87 | 42 min 16 sec |
| APG-BPDN | 0.5 | 1 | 6.83% | 52 | 12 min 32 sec |
| TVAL3 | 0.5 | 1 | 2.76% | 22 | 10 min 6 sec |
| proposed | 0.5 | 1 | 2.61% | 47 | 9 min 20 sec |

(D) HARVARD INDOOR

| algorithm | $r_p$ | $r_s$ | relative error | iterations | time |
|---|---|---|---|---|---|
| APG-BPDN | 0.2 | 0.1 | 15.26% | 200 | 81 min 45 sec |
| proposed | 0.2 | 0.1 | 10.63% | 150 | 58 min 25 sec |
| APG-BPDN | 0.2 | 1 | 8.35% | 54 | 12 min 37 sec |
| TVAL3 | 0.2 | 1 | 4.17% | 26 | 9 min 39 sec |
| proposed | 0.2 | 1 | 3.93% | 45 | 8 min 24 sec |
| APG-BPDN | 0.5 | 0.2 | 13.18% | 200 | 100 min 19 sec |
| proposed | 0.5 | 0.2 | 7.78% | 106 | 51 min 51 sec |
| APG-BPDN | 0.5 | 1 | 7.01% | 41 | 9 min 49 sec |
| TVAL3 | 0.5 | 1 | 3.57% | 29 | 12 min 41 sec |
| proposed | 0.5 | 1 | 3.30% | 39 | 7 min 45 sec |

(E) WASHINGTON DC MALL

| algorithm | $r_p$ | $r_s$ | relative error | iterations | time |
|---|---|---|---|---|---|
| APG-BPDN | 0.2 | 0.1 | 18.75% | 194 | 9 min 4 sec |
| proposed | 0.2 | 0.1 | 13.94% | 107 | 4 min 2 sec |
| APG-BPDN | 0.2 | 1 | 18.70% | 144 | 5 min 14 sec |
| TVAL3 | 0.2 | 1 | 13.73% | 57 | 3 min 10 sec |
| proposed | 0.2 | 1 | 13.67% | 59 | 1 min 47 sec |
| APG-BPDN | 0.5 | 0.2 | 12.71% | 172 | 9 min 24 sec |
| proposed | 0.5 | 0.2 | 7.92% | 91 | 4 min 30 sec |
| APG-BPDN | 0.5 | 1 | 12.43% | 130 | 4 min 39 sec |
| TVAL3 | 0.5 | 1 | 7.62% | 48 | 3 min 36 sec |
| proposed | 0.5 | 1 | 7.53% | 48 | 1 min 31 sec |

(F) INDIAN PINES

| algorithm | $r_p$ | $r_s$ | relative error | iterations | time |
|---|---|---|---|---|---|
| APG-BPDN | 0.2 | 0.1 | 6.67% | 179 | 24 min 6 sec |
| proposed | 0.2 | 0.1 | 4.62% | 116 | 15 min 17 sec |
| APG-BPDN | 0.2 | 1 | 6.16% | 73 | 6 min 41 sec |
| TVAL3 | 0.2 | 1 | 4.48% | 24 | 3 min 48 sec |
| proposed | 0.2 | 1 | 4.37% | 44 | 3 min 17 sec |
| APG-BPDN | 0.5 | 0.2 | 5.59% | 168 | 27 min 4 sec |
| proposed | 0.5 | 0.2 | 2.97% | 101 | 16 min 10 sec |
| APG-BPDN | 0.5 | 1 | 4.87% | 49 | 4 min 52 sec |
| TVAL3 | 0.5 | 1 | 2.77% | 22 | 3 min 47 sec |
| proposed | 0.5 | 1 | 2.62% | 33 | 2 min 34 sec |

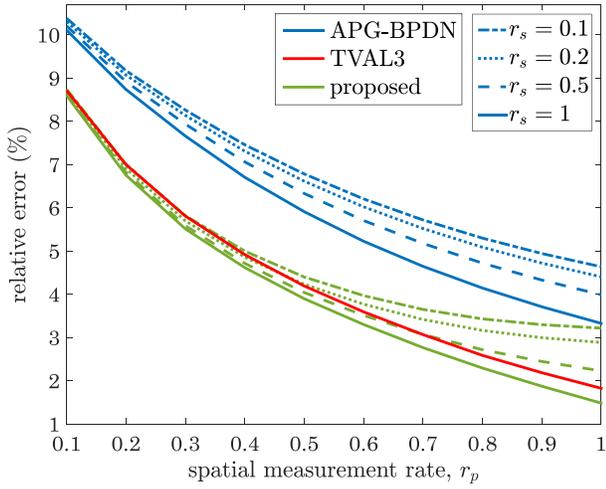

(a) Stanford Dish

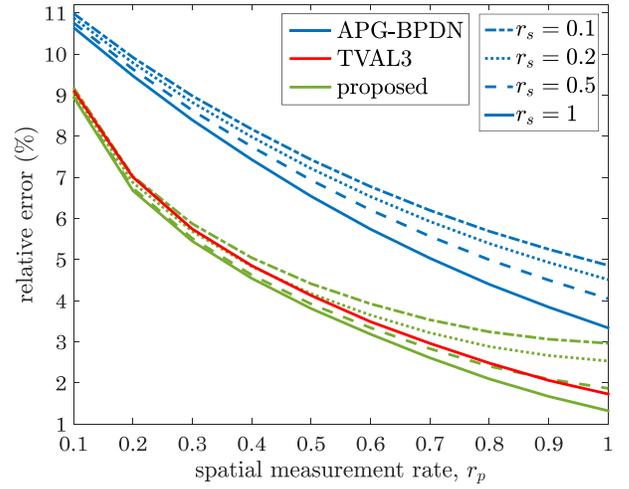

(b) San Francisco

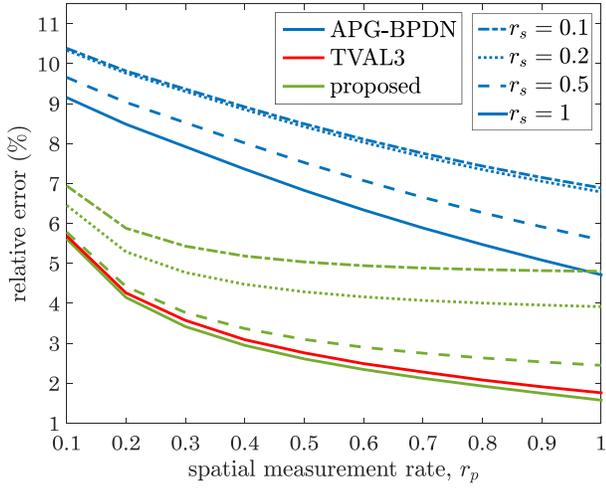

(c) Harvard Outdoor

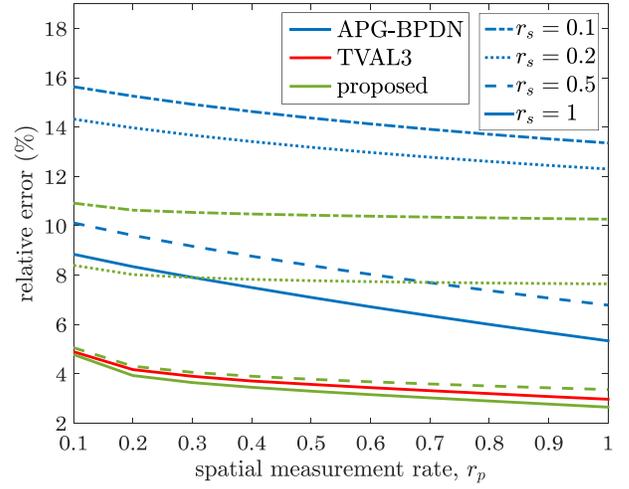

(d) Harvard Indoor

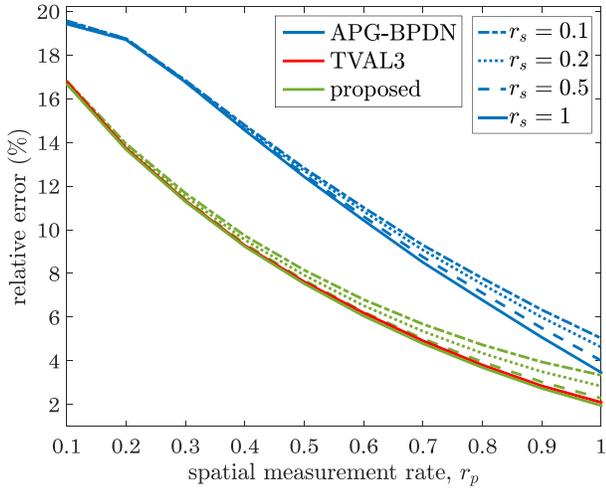

(e) Washington DC Mall

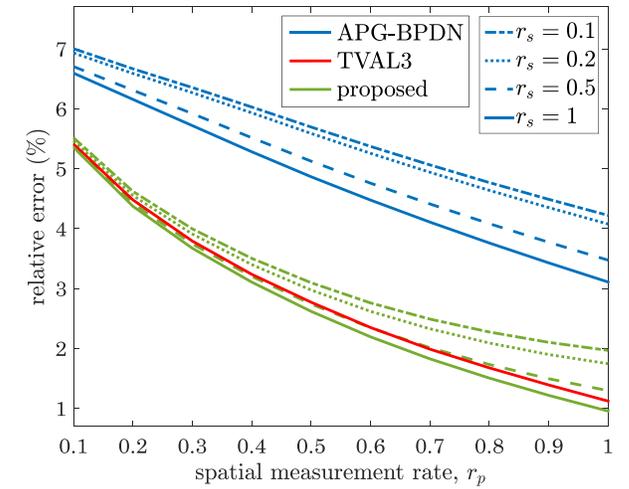

(f) Indian Pines

Fig. 2. Relative errors of recovering the considered hyperspectral images using the APG-BPDN, the TVAL3, and the proposed algorithms with different values of the spatial and spectral measurement rates.

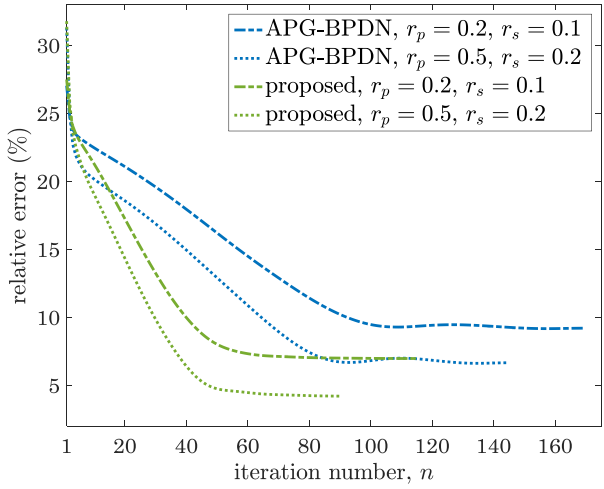

(a) Stanford Dish

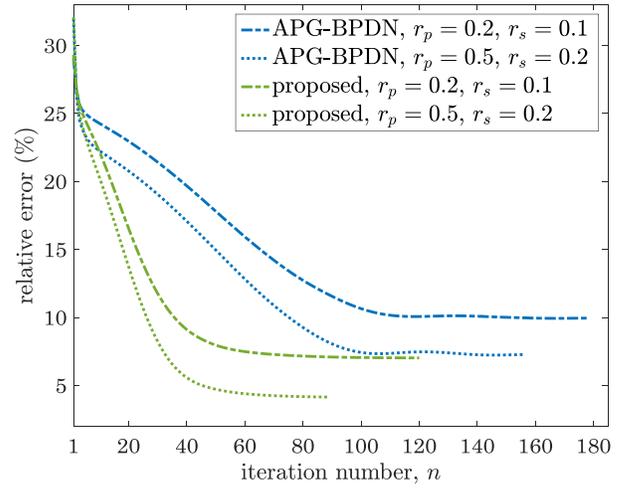

(b) San Francisco

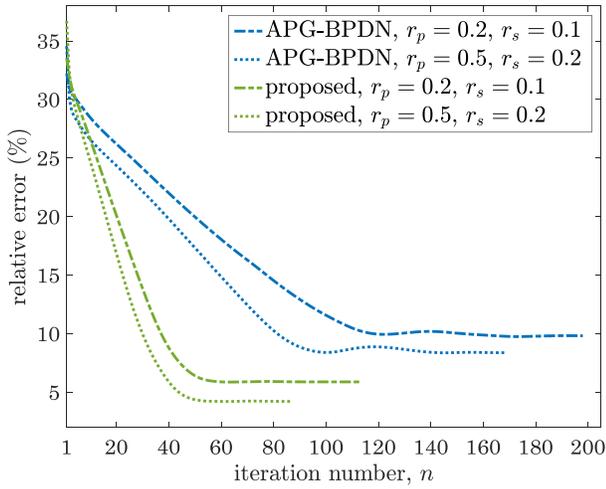

(c) Harvard Outdoor

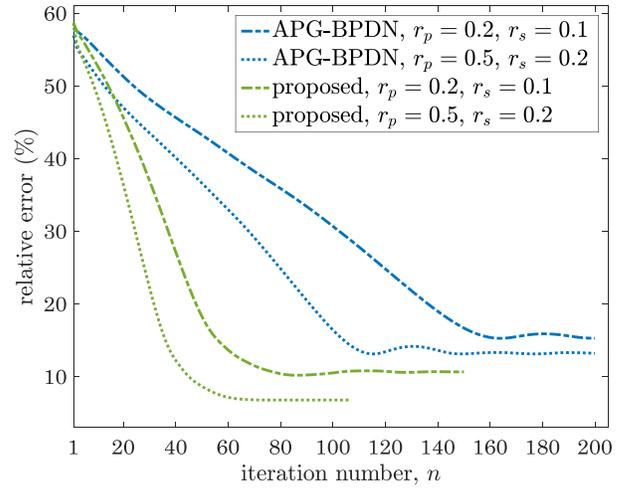

(d) Harvard Indoor

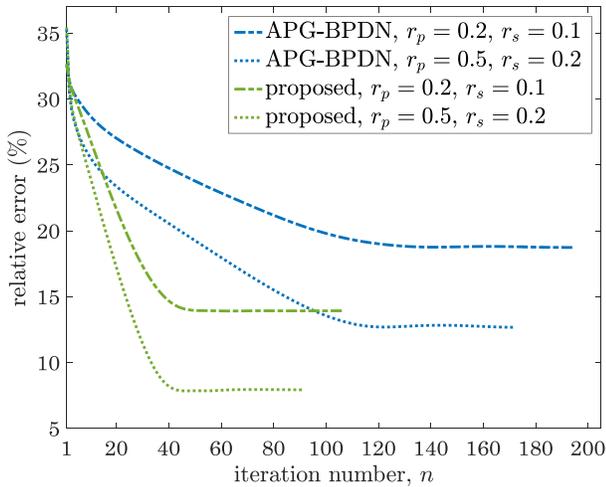

(e) Washington DC Mall

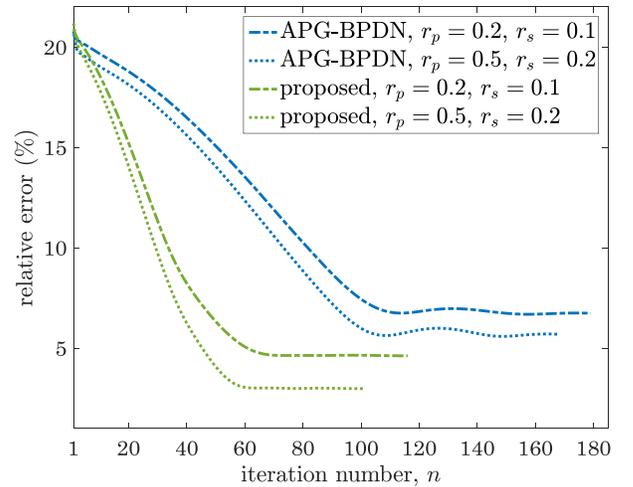

(f) Indian Pines

Fig. 3. Convergence curves of the APG-BPDN and the proposed algorithms for the considered hyperspectral images with different values of the spatial and spectral measurement rates.

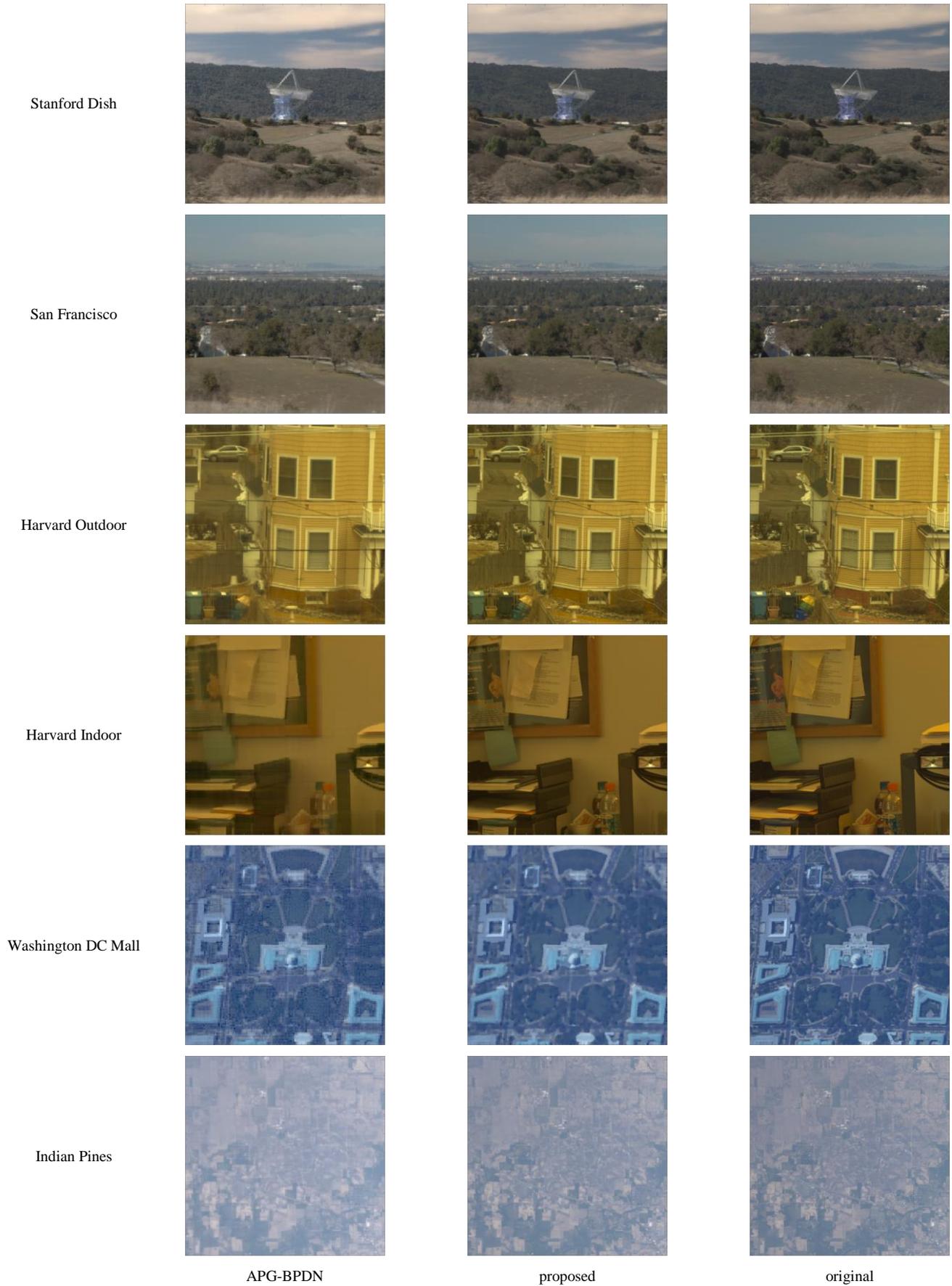

Fig. 4. RGB renderings of the original and recovered versions of the considered hyperspectral images. The recovery is performed using the APG-BPDN and the proposed algorithms when $r_p = 0.2$ and $r_s = 0.1$.

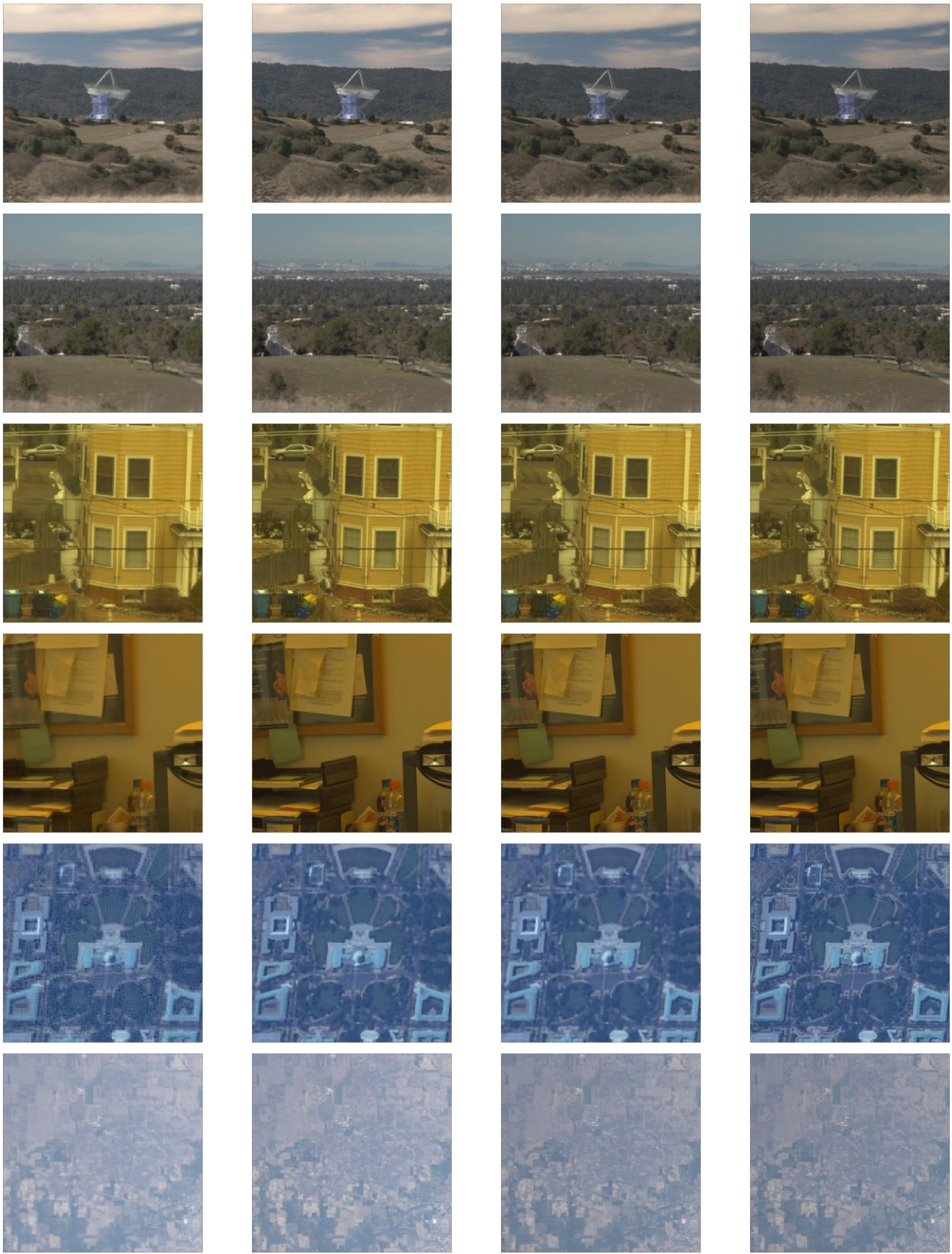

APG-BPDN      TVAL3      proposed      original

Fig. 5. RGB renderings of the original and recovered versions of the considered hyperspectral images. The recovery is performed using the APG-BPDN, the TVAL3, and the proposed algorithms when $r_p = 0.2$ and $r_s = 1$.

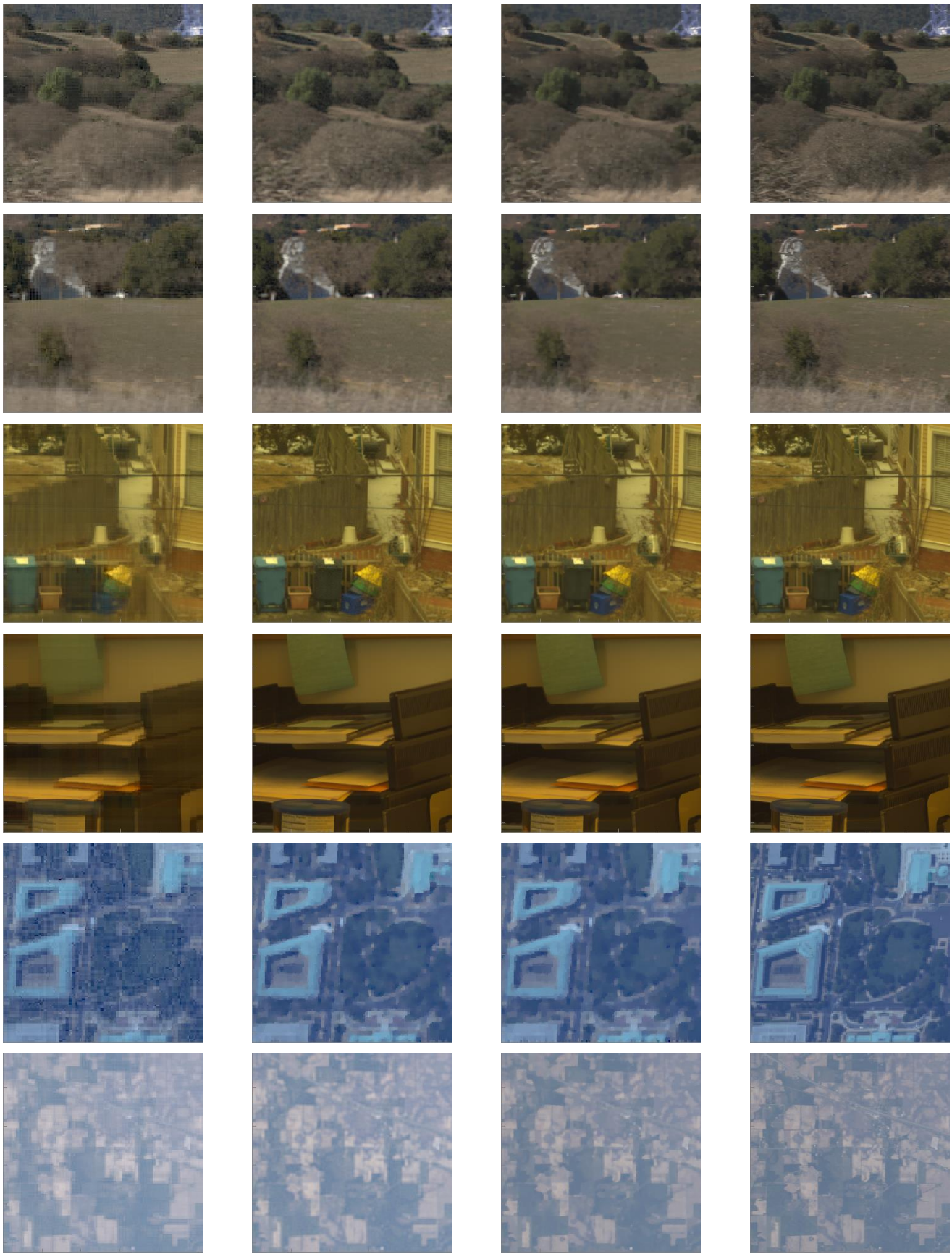

APG-BPDN      TVAL3      proposed      original

Fig. 6. Images in Fig. 5 zoomed to their bottom left quarter.

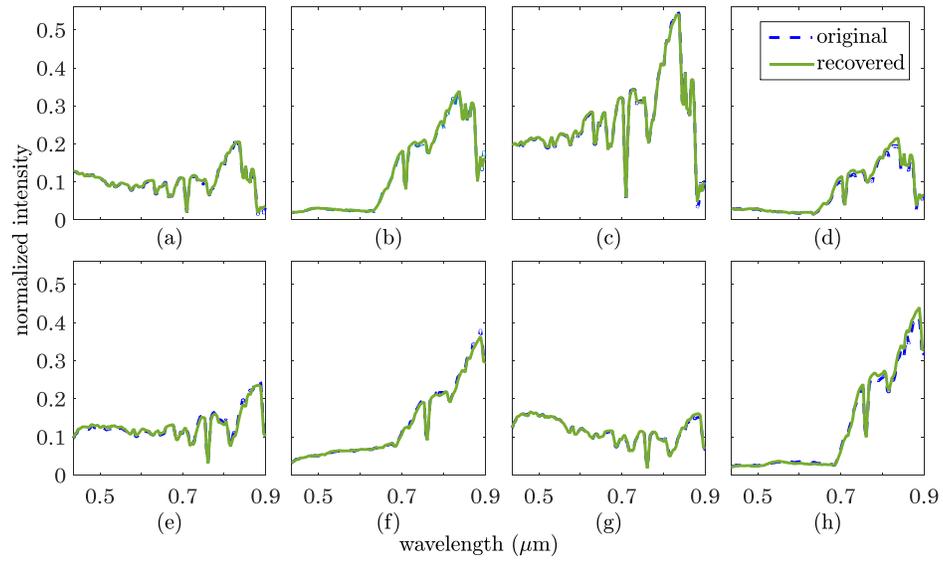

Fig. 7.  Original and recovered spectra of four pixels randomly selected from each of the Stanford Dish (a)-(d) and San Francisco (e)-(h) hyperspectral images. The recovery is performed using the proposed algorithm when $r_p = 0.2$ and $r_s = 0.1$.